\documentclass{article}
\usepackage{ijcai17}
\usepackage{footmisc}
\usepackage{times}

\usepackage{hyperref}
\usepackage{url}

\usepackage{caption}
\usepackage{subcaption}

\usepackage{graphicx}
\usepackage{amsmath,amssymb} 
\usepackage{multirow}
\usepackage{algorithm}
\usepackage{algpseudocode}  
\usepackage{amsmath} 
\usepackage{color}

\title{Reinforcement Learning for Learning Rate Control}

\author{Chang Xu$^{1}$,Tao Qin$^{2}$,Gang Wang$^{1}$,Tie-Yan Liu$^{2}$\\
	$^{1}$College of Computer and Control Engineering, Nankai University, Tianjin, China\\ $^{2}$Microsoft Research, Beijing, China\\
	$^{1}$\{changxu, wgzwp\}@nbjl.nankai.edu.cn, $^{2}$\{taoqin, tie-yan Liu\}@microsoft.com}

\begin{document}

\maketitle

\begin{abstract}
	
	Stochastic gradient descent (SGD), which updates the model parameters by adding a local gradient times a learning rate at each step, is widely used in model training of machine learning algorithms such as neural networks. It is observed that the models trained by SGD are sensitive to learning rates and good learning rates are problem specific.
	We propose an algorithm to automatically learn learning rates using neural network based actor-critic methods from deep reinforcement learning (RL).
	In particular, we train a policy network called actor to decide the learning rate at each step during training, and a value network called critic to give feedback about quality of the decision (e.g., the goodness of the learning rate outputted by the actor) that the actor made. The introduction of auxiliary actor and critic networks helps the main network achieve better performance.
	Experiments on different datasets and network architectures show that our approach leads to better convergence of SGD than human-designed competitors.
	
\end{abstract}

\section{Introduction}

While facing large scale of training data, stochastic learning such as stochastic gradient descent (SGD) is usually much faster than batch learning and often results in better models.  An observation for SGD methods is that their performances are highly sensitive to the choice of learning rate ~\cite{lecun2012efficient}. Clearly, setting a static learning rate for the whole training process is insufficient, since intuitively the learning rate should decrease when the model becomes more and more close to a (local) optimum as the training goes on over time ~\cite{maclaurin2015gradient}. Although there are some empirical suggestions to guide how to adjust the learning rate over time in training, it is still a difficult task to find a good policy to adjust the learning rate, given that good policies are problem specific and depend on implementation details of a machine learning algorithm.  One usually needs to try many times and adjust the learning rate manually to accumulate knowledge about the problem. However, human involvement often needs domain knowledge about the target problems, which is inefficient and difficult to scale up to different problems. Thus, a natural question arises: can we learn to adjust the learning rate? This is exactly the focus of this work and we aim to learn learning rates for SGD based machine learning (ML) algorithms without human-designed rules or hand-crafted features. 

By examining the current practice of learning rate control/adjustment, we have two observations. First, learning rate control is a sequential decision process: We set an initial learning rate at the beginning, and then at each step we decide whether to change the learning rate and how to change it, based on the current model and loss, training data at hand, and maybe history of the training process. As suggested in~\cite{orr2003neural}, one well-principled method for estimating the ideal learning rate that is to decrease the learning rate when the weight vector oscillates, and increase it when the weight vector follows a relatively steady direction.  Second, although at each step some immediate reward (e.g., the loss decrement) can be obtained by taking actions, we care more about the performance of the final model found by the ML algorithm. Consider two different learning rate control policies: the first one leads to fast loss decrease at the beginning but gets saturated and stuck in a local minimum quickly, while the second one starts with slower loss decrease but results in much smaller final loss. Obviously, the second policy is better. That is, we prefer long-term rewards over short-term rewards.

Combining the two observations, it is not difficult to see that the problem of finding a good policy to control/adjust learning rate falls into the scope of reinforcement learning (RL) ~\cite{sutton1998reinforcement}. Inspired by the recent success of RL for sequential decision problems, in this work, we leverage RL techniques and try to learn learning rate for SGD based methods.

We propose an algorithm to learn learning rate within the actor-critic framework ~\cite{sutton1984temporal,sutton1999policy,barto1983neuronlike,lever2014deterministic}, which is widely used in RL. An actor network is trained to take an action that decides the learning rate for current step, and a critic network is trained to give feedback to the actor network about long-term performance and help the actor network to adjust itself so as to perform better in the future. To reduce oscillation during training, we take gradient disagreement among training samples into account. By feeding different training samples to the actor network and the critic network, learning rate is encouraged to be small when gradients oscillate, which is consistent with the suggestion for ideal learning rate strategy in ~\cite{orr2003neural}. A series of experiments on different datasets and network architectures validate the effectiveness of our proposed algorithm for learning rate control.

The main contributions of this paper include:
\begin{itemize}
	\setlength{\itemsep}{0pt}
	\setlength{\parsep}{0pt}
	\setlength{\parskip}{0pt}
	\item[\textbullet] We propose to use actor-critic based auxiliary networks to learn and control the learning rate for ML algorithms, so that the ML algorithm can achieve better convergence.
	
	\item[\textbullet] 
	Long-term rewards are exploited in our approach rather than only immediate rewards (e.g., the decrease of loss for one step). The expected total decrease of loss in future steps is modeled by the critic network, so that the actor can make far-sighted decision for learning rate.
\end{itemize}

\section{Related Work}
We review some related work in this section.

\subsection{Improved Gradient Methods}

Our focus is to improve gradient based ML algorithm through automatic learning of learning rate. Different approaches have been proposed to improve gradient methods, especially for deep neural networks.

Since SGD solely rely on a given example (or a mini-batch of examples) to compare gradient, its model update at each step tends to be unstable and it takes many steps to converge. To solve this problem, momentum SGD~\cite{jacobs1988increased} is proposed to accelerate SGD by using recent gradients. 
RMSprop ~\cite{tieleman2012lecture} utilizes the magnitude of recent gradients to normalize the gradients. It always keeps a moving average over the root mean squared gradients, by which it divides the current gradient. 
Adagrad ~\cite{duchi2011adaptive} adapts component-wise learning rates, and performs larger updates for infrequent and smaller updates for frequent parameters.
Adadelta ~\cite{zeiler2012adadelta} extends Adagrad by reducing its aggressive, monotonically decreasing learning rate. Instead of accumulating all past squared gradients, Adadelta restricts the window of accumulated past gradients to some fixed size. 
Adam ~\cite{kingma2014adam} computes component-wise learning rates using the estimates of first and second moments of the gradients, which combines the advantages of AdaGrad and RMSProp.

\cite{senior2013empirical,sutton1992adapting,darken1990fast} focus on predefining update rules to adjust learning rates during training. A limitation of these methods is that they have additional free parameters which need to be set manually. \cite{schaul2013no} proposes a method to choose good learning rate for SGD, which relies on the square norm of the expectation of the gradient, and the expectation of the square norm of the gradient. The method is much more constrained than ours and several assumption should be met. 
Another recent work \cite{daniel2016learning} investigates several hand-tuned features and uses the Relative Entropy Policy Search method as controller to select step size for SGD and RMSprop.

\subsection{Reinforcement Learning}
Since our proposed algorithm is based on RL techniques, here we give a very brief introduction to RL, which will ease the description of our algorithm in next section.

Reinforcement learning ~\cite{sutton1988learning} is concerned with how an agent acts in a stochastic environment by sequentially choosing actions over a sequence of time steps, in order to maximize a cumulative reward. In RL, a state $s^t$ encodes the agent's observation about the environment at a time step $t$, and a policy function $\pi(s^t)$ determines how the agent behaves (e.g., which action to take) at state $s^t$. An action-value function (or, Q function) $Q_{\pi}(s^t,a^t)$ is usually used to denote the cumulative reward of taking action $a^t$ at state $s^t$ and then following policy $\pi$ afterwards.   

Many RL algorithms have been proposed~\cite{sutton1998reinforcement,watkins1992q}, and many RL algorithms ~\cite{sutton1984temporal,sutton1999policy,barto1983neuronlike,lever2014deterministic} can be described under the actor-critic framework.  An actor-critic algorithm learns the policy function and the value function simultaneously and interactively. The policy structure is known as the actor, and is used to select actions; the estimated value function is known as the critic, and it criticizes the actions made by the actor.

Recently, deep reinforcement learning, which uses deep neural networks to approximate/represent the policy function and/or the value function, have shown promise
in various domains, including Atari games ~\cite{mnih2015human}, Go ~\cite{silver2016mastering}, machine translation ~\cite{bahdanau2016actor}, image recognition ~\cite{xu2015show}, etc. 

\section{Method}
In this section, we present an actor-critic algorithm that can automate the learning rate control for SGD based machine learning algorithms. 
\begin{figure}[ht]
	\centering
	\includegraphics[page=1, width=0.9\linewidth]{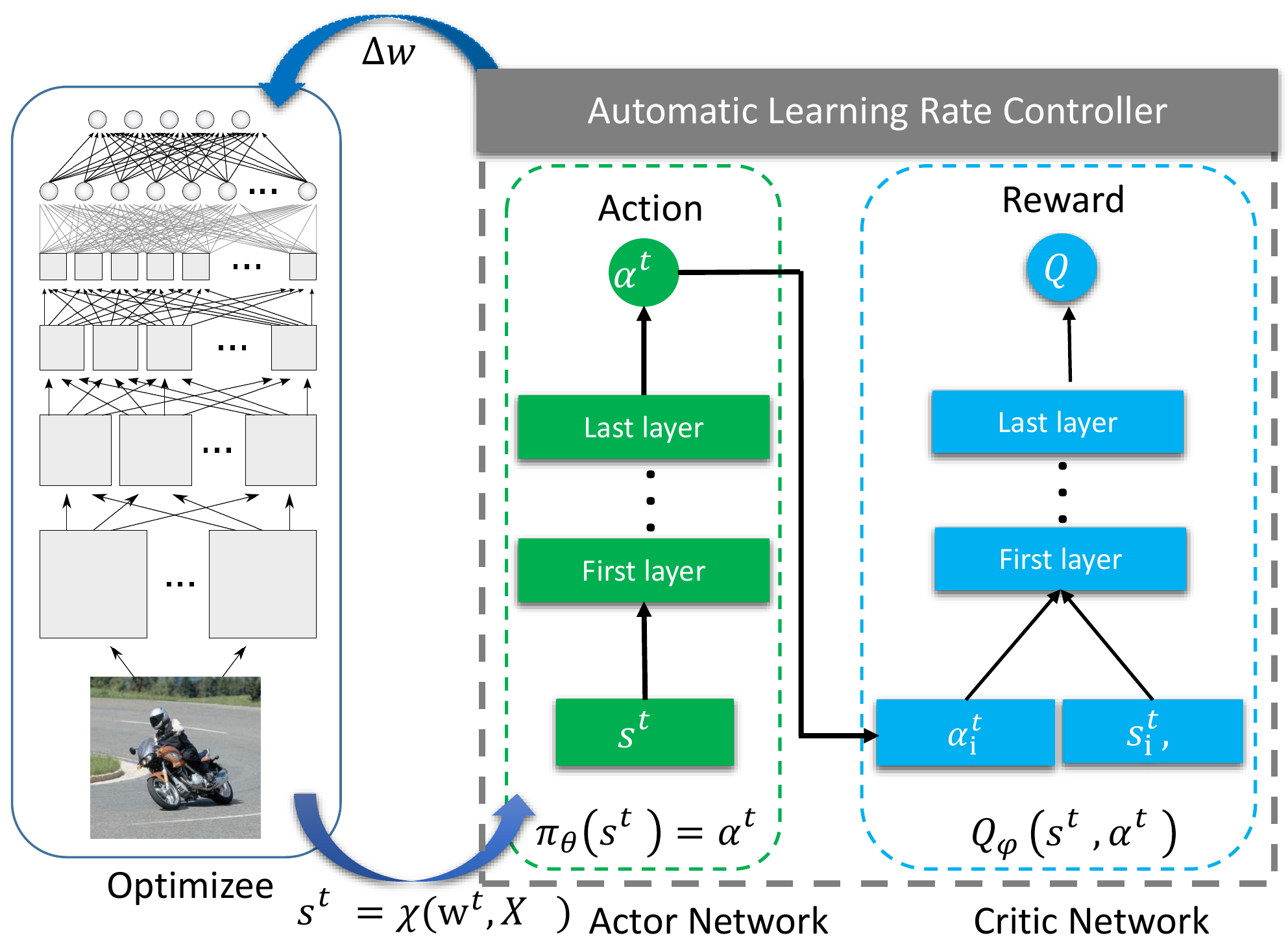}
	\caption{The framework of our proposed automatic learning rate controller.}
	\label{fig_framework}
\end{figure}

Many machine learning tasks need to train a model with parameters $\omega$ by minimizing a loss function $f$ defined over a set $X$ of training examples:
\begin{equation}
	\omega^* = \mathop{\arg\min }_{\omega}f_\omega(X).
\end{equation} 
A standard approach for the loss function minimization is gradient descent, which sequentially updates the parameters using gradients step by step:  
\begin{equation} \label{omega_update}
	\omega^{t+1} = \omega^t - a^t\nabla f^t,
\end{equation}
where $a^t$ is the learning rate at step $t$, and $\nabla f^t$ is the local gradient of $f$ at $\omega^t$. Here one step can be the whole batch of all the training data, a mini batch of tens/hundreds of examples, or a random sample. 

It is observed that the performance of SGD based methods is quite sensitive to the choice of $a^t$ for non-convex loss function $f$. Unfortunately, $f$ is usually non-convex with respect to the parameters $w$ in many ML algorithms, especially for deep neural networks.  We aim to learn a learning rate controller using RL techniques that can automatically control $a^t$.

Figure \ref{fig_framework} illustrates our automatic learning rate controller, which adopts the actor-critic framework in RL. The basic idea is that at each step, given the current model $\omega^t$ and training sample $x$, an actor network is used to take an action (the learning rate $a^t$, and it will be used to update the model $\omega^t$), and a critic network is used to estimate the goodness of the action. The actor network will be updated using the estimated goodness of $a^t$, and the critic network will be updated by minimizing temporal difference (TD) ~\cite{sutton1990time} error.

We describe the details of our algorithm in the following subsections.

\algnewcommand{\IIf}[1]{\State\algorithmicif\ #1\ \algorithmicthen}
\algnewcommand{\EndIIf}{\unskip\ \algorithmicend\ \algorithmicif}
\begin{algorithm}[!ht]  
	\caption{Actor-Critic Algorithm for Learning Rate Learning}  
	\label{alg:Framwork}  
	\begin{algorithmic}[1]  
		\Require  
		Training steps $ T $ ;
		training set $ X $;
		loss function $f$;
		state function $\chi$;
		discount factor: $\gamma$ ;
		mini-batch size $m_\theta$ of actor network;
		mini-batch size $m_\varphi$ of critic network;
		reset frequency of the model $ e $
		\State Initialize model parameters $\omega$ as $\omega_0$, policy parameters $\theta$ of the actor network as $\theta_0$, and value parameters $\varphi$ of the critic network as $\varphi_0$.
		\For{$t=1,...,T$}
		\State Sample $x_{i}\in X, i \in {1,...,N}.$
		\State Extract state vector: $s^t_i = \chi(\omega^t, x_{i})$.
		\State $ // $Actor network selects an action.
		\State Computes learning rate $a^t_i = \pi_\theta(s^t_i)$.
		\State $ // $Update model parameters $\omega$.
		\State Compute $\nabla f^t(x_{i})$.
		\State Update $\omega$: $\omega^{t+1}=\omega^t - a^t_i \nabla f^t(x_{i})$.
		\State $ // $Update critic network by minimizing square error between estimation and label.
		\State $r^t = f^t(x_i) - f^{t+1}(x_i)$
		\State Extract state vector: $s^{t+1}_i = \chi(\omega^{t+1}, x_i)$
		\State Compute $Q_\varphi(s^{t+1}_i,\pi_\theta(s^{t+1}_i)), Q_\varphi(s^{t}_i,a^{t}_i)$
		\State Compute $\delta^t$ according to Equation \ref{compute_delta}:
		
		$\delta^t =  r^t + \gamma Q_\varphi(s^{t+1}_i, \pi_\theta(s^{t+1}_i))- Q_\varphi(s^t_i, a^t_i)$
		\State Compute the gradients of critic network according to Equation \ref{update_varphi} :
		
		$\nabla\varphi^t =\delta^t\nabla_\varphi Q_\varphi(s^t_i, a^t_i)$
		\If {$t \bmod m_\varphi = 0$}
		\State Update $\varphi$ by $\nabla\varphi=\frac{1}{m_\varphi}\sum_{u=0}^{m_\varphi - 1}\nabla\varphi^{t-u}$
		\EndIf
		\State $ // $ Update actor network
		\State Sample  $ x_{j} \in X , j \in {1,...,N}, j \neq i$.
		\State Extract state vector: $s^{t+1}_j = \chi(\omega^{t+1}, x_{j})$.
		\State Compute $a^{t+1}_j = \pi_\theta(s^{t+1}_j)$.
		\State Compute the gradients of actor network according to Equation \ref{update_theta}:
		
		$\nabla\theta^t = \nabla_\theta \pi_\theta(s^{t+1}_j)\nabla_a Q_\varphi(s^{t+1}_j,a^{t+1}_j)|_{a=\pi_\theta(s)}$
		\If {$t \bmod m_\theta = 0$}
		\State Update $\theta$ by $\nabla\theta=\frac{1}{m_\theta}\sum_{u=0}^{m_\theta - 1}\nabla\theta^{t-u}$
		\EndIf
		
		\IIf{$t \bmod e = 0$} set $\omega_{t+1} = \omega_0$ \EndIIf
		\EndFor\\
		\Return $ \omega, \theta, \varphi$;
		
	\end{algorithmic}  
\end{algorithm}

\subsection{Actor Network}
The actor network, which is called policy network in RL, plays the key role in our algorithm: it determines the learning rate control policy for the primary ML algorithm\footnote{Here we have two learning algorithms. We call the one with learning rate to adjust as the primary ML algorithm, and the other one which optimizes the learning rate of the primary one as the secondary ML algorithm.} based on the current model, training data, and maybe historical information during the training process.

Note that $\omega^t$ could be of huge dimensions, e.g., one widely used image recognition model VGGNet ~\cite{Simonyan14c} has more than 140 million parameters. If the actor network takes all of those parameters as the inputs, its computational complexity would dominate the complexity of the primary algorithm, which is unfordable. 
Therefore, we propose to use a function $\chi (\cdot)$ to process and yield a compact vector $s^t$ as the input of the actor network. Following the practice in RL, we call $\chi (\cdot)$ the state function, which takes $\omega^t$ and the training data $x$ as inputs:
\begin{equation}
	s^t = \chi(\omega^t, X).
\end{equation}
Then the actor network $\pi_\theta(\cdot)$ parameterized by $\theta$ yields an action $ a^t $:
\begin{equation}
	\pi_\theta(s^t) = a^t,
\end{equation} 
where the action $a^t \in \mathbb{R}$ is a continuous value. 
When $a^t$ is determined, we update the model of the primary algorithm by Equation \ref{omega_update}.

Note that the actor network has its own parameters and we need to learn them to output a good action. To learn the actor network, we need to know how to evaluate the goodness of an actor network. The critic network exactly plays this role.

\begin{figure*}[!ht]
	\centering
	\begin{minipage}[t]{0.5\linewidth}
		\centering
		\includegraphics[page=24, width=0.8\linewidth]{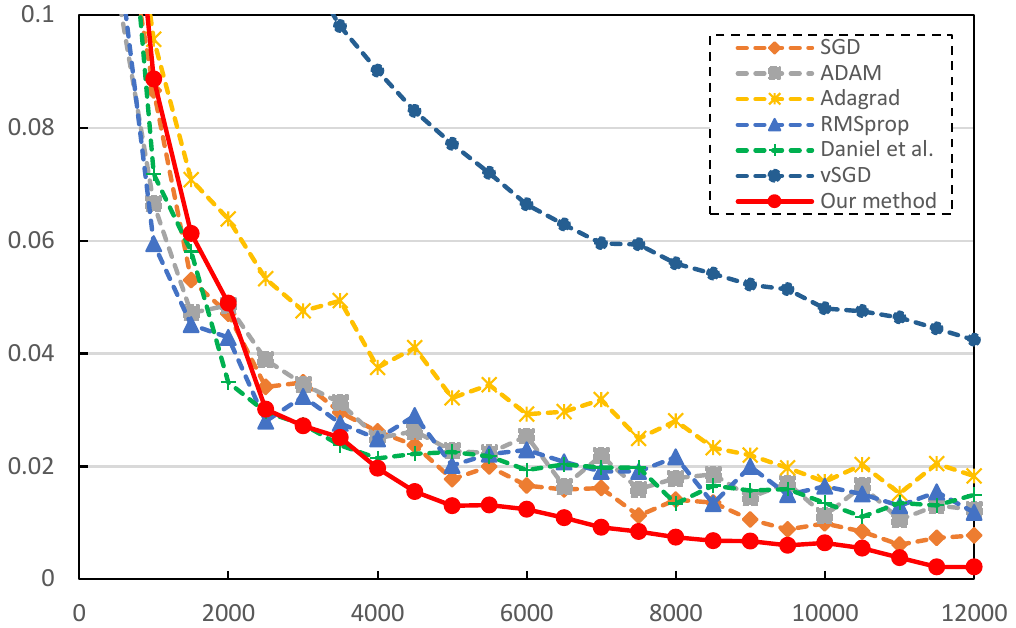}
		
		(a)
	\end{minipage}%
	\begin{minipage}[t]{0.5\linewidth}
		\centering
		\includegraphics[page=25, width=0.8\linewidth]{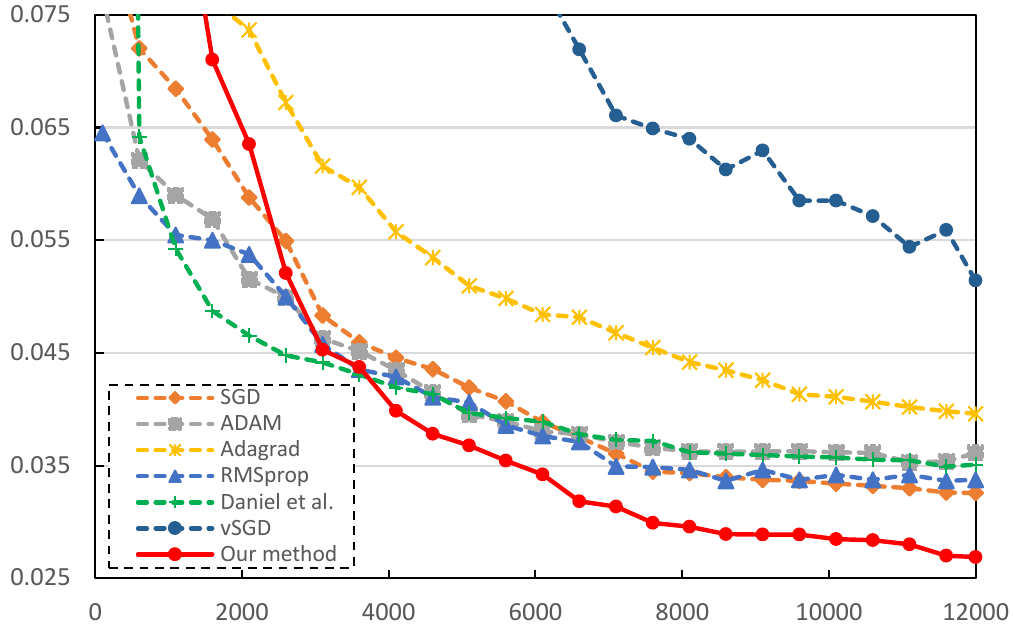}
		
		(b)
	\end{minipage}
	\vspace*{-2mm}
	\caption{Results on MNIST. (a) Training loss. (b) Test loss. The x-axis represents the number of mini batches. The y-axis represents loss value.}
	\vspace*{-2mm}
	\label{exp_mnist}
	
\end{figure*}

\subsection{Critic Network}
Recall that our goal is to find a good policy for learning rate control to ensure that a good model can be learnt eventually by the primary ML algorithm. For this purpose, the actor network needs to output a good action $a^t$ at state $s^t$ so that finally a low training loss $f(\cdot)$ can be achieved. In RL, the Q function $Q_\pi(s, a)$ is often used to denote the long term reward of the state-action pair $s,a$ while following the policy $\pi$ to take future actions. In our problem, $Q_\pi(s^t, a^t)$ indicates the accumulative decrement of training loss starting from step $t$. We define the immediate reward at step $t$ as the one step loss decrement:
\begin{equation}
	r^t = f^t - f^{t+1}.
\end{equation}
The accumulative value $R_\pi^t $ of policy $\pi$ at step $t$ is the total discounted reward from step $t$: $$R_\pi^t = \Sigma^T_{k=t} \gamma^{k-t}r(s^k,a^k),$$ where  $\gamma\in (0,1]$ is the discount factor. 

Considering that both the states and actions are uncountable in our problem, the critic network uses a parametric function $Q_\varphi(s,a)$ with parameters $\varphi$ to approximate the Q value function $Q_\pi(s, a)$. 


\subsection{Training of Actor and Critic Networks}
The critic network has its own parameters $\varphi$, which is updated at each step using TD learning. More precisely, the critic is trained by minimizing the square error between the estimation $Q_\varphi(s^{t}, a^{t})$ and the target $y^t$:
\begin{equation}
	y^t = r^t + \gamma Q_\varphi(s^{t+1}, a^{t+1}).
\end{equation}
The TD error is defined as: 
\begin{equation} \label{compute_delta}
	\begin{array}{rcl}
		\delta^t & = & y^t - Q_\varphi(s^t, a^t) \\
		& = & r^t + \gamma Q_\varphi(s^{t+1}, \pi_\theta(s^{t+1}))- Q_\varphi(s^t, a^t)
	\end{array}	
\end{equation} 
The weight update rule follows the on-policy deterministic actor-critic algorithm. The gradients of critic network are:
\begin{equation}\label{update_varphi}
	\nabla\varphi = \delta^t\nabla_\varphi Q_\varphi(s^t, a^t),
\end{equation}

The policy parameters $\theta$ of the actor network is updated by ensuring that it can output the action with the largest Q value at state $s^t$, i.e., $a^*=\arg\max_{a}Q_\varphi(s^{t}, a)$. Mathematically, 
\begin{equation}\label{update_theta}
	\nabla\theta = \nabla_\theta \pi_\theta(s^{t+1})\nabla_a Q_\varphi(s^{t+1},a^{t+1})|_{a=\pi_\theta(s)}.
\end{equation}

\subsection{The Algorithm} 
\label{overfitting_section}
The overall algorithm for learning rate learning is shown in Algorithm 1. In each step, we sample an example (Line 3), extract the current state vector (Line 4), compute the learning rate using the actor network (Line 6), update the model (Lines 8-9), compute TD error (Lines 11-15), update the critic network (Line 16-18), and sample another example (Line 20) to update the actor network (Line 21-26). We would like to make some discussions about the algorithm.

First, in the current algorithm, for simplicity, we consider using only one example for model update. It is easy to generalize to a mini batch of random examples.

Second, one may notice that we use one example (e.g., $x_i$) for model and the critic network update, but a different example (e.g., $x_j$) for the actor network update. 

Doing so we can reduce oscillation during training. Suppose that the gradient direction of current example (or mini batch of examples) is quite different from others in this stage of training process. Intuitively at this step the model will be changed a lot to fit the example, consequently resulting in oscillation of the training, as shown in our experiments. As aforementioned, one principle for ideal learning rate control is to decrease it when the gradient vector oscillates, and increase it when the gradient vector follows a relatively steady direction. Therefore, we try to alleviate the problem by controlling learning rate according to gradient disagreement.

By feeding different examples to the actor and critic networks, it is very likely the critic network will find that the gradient direction of the example fed into the actor network is inconsistent with its own training example and thus criticize the large learning rate suggested by the actor network. More precisely, the update of $\omega$ is based on $x_i$ and the learning rate suggested by the actor network, while the training target of the actor network is to maximize the output of the critic network on $x_j$. If there is big gradient disagreement between $x_i$ and $x_j$, the update of $\omega$, which is affected by actor's decision, would cause the critic's output on $x_j$ to be small. To compensate this effect, the actor network is forced to predict a small learning rate for big gradient disagreement in this situation.

\section{Experiments}
We conducted a set of experiments to test the performance of our learning rate learning algorithm and compared with several baseline methods. We report the experimental results in this section.

\begin{figure*}[!ht]
	\centering
	\begin{minipage}[t]{0.5\linewidth}
		\centering
		\includegraphics[page=26, width=0.82\linewidth]{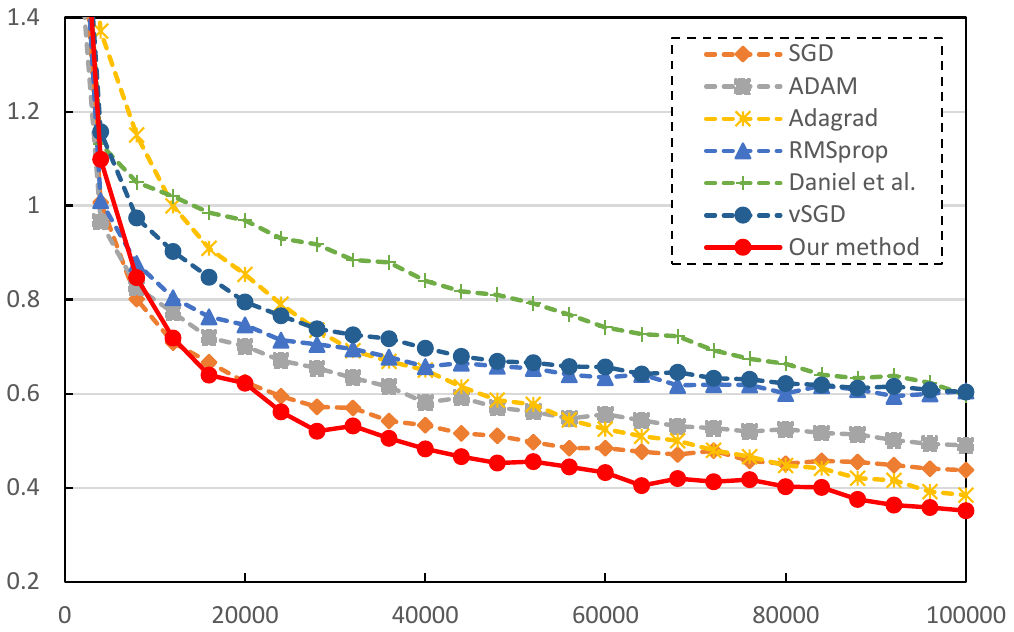}
		
		(a)
	\end{minipage}%
	\begin{minipage}[t]{0.5\linewidth}
		\centering
		\includegraphics[page=27, width=0.82\linewidth]{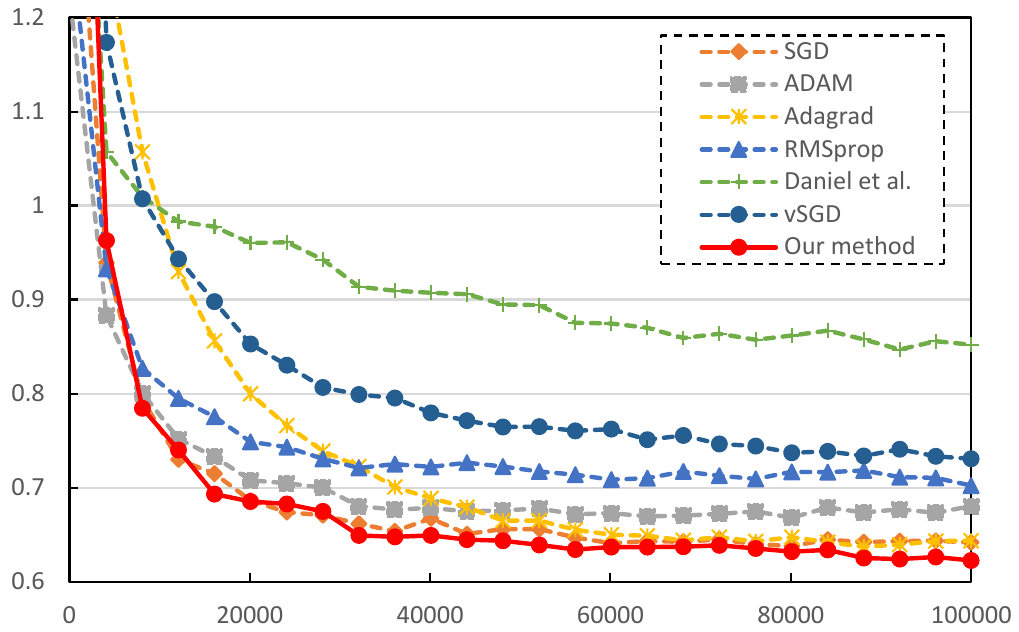}
		
		(b)
	\end{minipage}
	\vspace*{-2mm}
	\caption{ Results on CIFAR10. (a) Training loss. (b) Test loss. The x-axis is the number of mini batches. The y-axis represents loss value.}
	\label{exp_cifar10}
	\vspace{-1.5em}
\end{figure*}
\subsection{Experimental Setup}

We specified our actor-critic algorithm in experiments as follows. Given that stochastic mini-batch training is a common practice in deep learning, the actor-critic algorithm also operated on mini-batches, i.e., each step is a mini batch in our experiments. The state $s^t = \chi(\omega^t, X_i)$ is defined as the average loss of learning model $\omega^t$ on the input mini batch $X_i$.The actor network is specified as a two-layer long short-term memory (LSTM) network
with 20 units in each layer, 
considering that a good learning rate for step $t$ depends on and correlates with the learning rates at previous steps while LSTM is well suited to model sequences with long-distance dependence. 
The critic network is specified as a simple neural network with one hidden layer and 10 hidden units. Adam with the default setting in toolbox is used to train the learning rate learner in all the experiments.

We compared our method with several mainstream SGD algorithms, including SGD, Adam~\cite{kingma2014adam}, Adagrad~\cite{duchi2011adaptive} and RMSprop~\cite{tieleman2012lecture}. We also compare our method with a recent work by \cite{daniel2016learning}\footnote{Thank the authors for providing the source code.\label{note}}. This work identifies several hand-designed features and use an RL method (Relative Entropy Policy Search) to learn a learning rate controller. Another baseline method is ``vSGD" \cite{schaul2013no}\footref{note}, which automatically adjusts learning rates to minimize the expected error. It tries to compute learning rate at each update by optimizing the expected loss after the next update according to (1) the square norm of the expectation of the gradient and (2) the expectation of the square norm of the gradient.

\subsection{Experimental Results}
To verify the effectiveness of our method on different datasets and model structures, experiments are conducted on two widely used image classification datasets:  MNIST ~\cite{lecun1998gradient} and CIFAR-10 ~\cite{krizhevsky2009learning}. For simplicity, the primary ML algorithm adopted the CNN models and settings from tensorflow~\cite{abadi2015tensorflow} tutorial, whose source code can be found at ~\cite{TensorflowExamples}
For each of these algorithms and each dataset,  
we tried the following learning rates $10^{-4},10^{-3},...,10^0$. We 
report the best performance of these algorithms
over those learning rates.
If an algorithm needs some other parameters to set, such as decay coefficients for Adam, we used the default setting in the toolbox. 
For each benchmark and our proposed method, five independent runs are averaged and reported in all of the following experiments. We trained all the baseline models until convergence.

\subsubsection{Results on MNIST}

MNIST is a dataset for handwritten digit classification task. Each example in the dataset is a $ 28\times28 $ black and white image containing a digit in $\{0,1,\cdots, 9\}$. The CNN model used in the primary ML algorithm is consist of two convolutional layers, each followed by a pooling layer, and finally a fully connected layer. 
There are 60,000 training images and 10,000 test images in this dataset. We scaled the pixel values to the [0,1] range before inputting to all the algorithms. Each mini batch contains 50 randomly sampled images.

Figure \ref{exp_mnist} shows the results of our actor-critic algorithm and the baseline methods, including the curves of training loss and test loss. We have the following observations.

\begin{itemize}
	
	\item Although the loss of our algorithm does not decrease very fast at the beginning, our algorithm achieves the best performance at the end. One may expect that our algorithm should have significantly faster convergence speed from the beginning considering that our algorithm learns both the learning rate and the CNN model, while most of the baseline methods only learn the CNN model and choose the learning rate per some predefine rules. However, this is not the case. Since our method targets at future long-term rewards rather than immediate rewards, it can make far-sighted decision and lead to better performance in long term. 
	\item The loss curves of our approach is more smooth and stable than others. That is because we carefully design the algorithm and feed different samples to the actor network and critic network. As discussed in Section\ref{overfitting_section}, doing so we can reduce oscillation during training.
	
\end{itemize}

\begin{figure*}[!ht]
	\centering
	\begin{minipage}[t]{0.3\linewidth}
		\centering
		\includegraphics[page=17, width=1\linewidth]{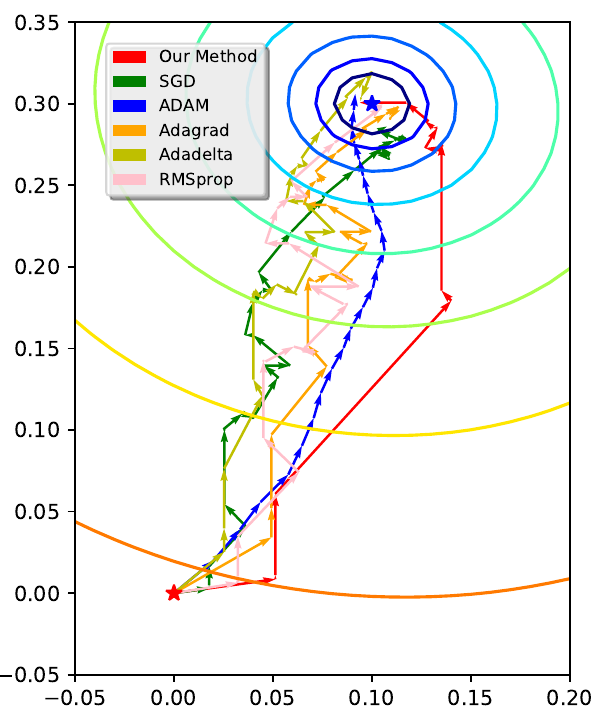}
		
		(a)
	\end{minipage}
	\begin{minipage}[t]{0.3\linewidth}
		\centering
		\includegraphics[page=18, width=1\linewidth]{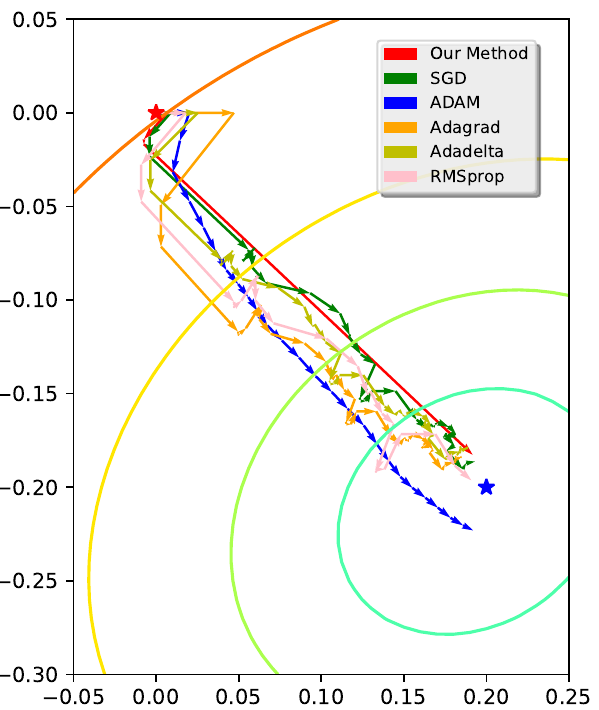}
		
		(b)
	\end{minipage}
	\begin{minipage}[t]{0.3\linewidth}
		\centering
		\includegraphics[page=19, width=1\linewidth]{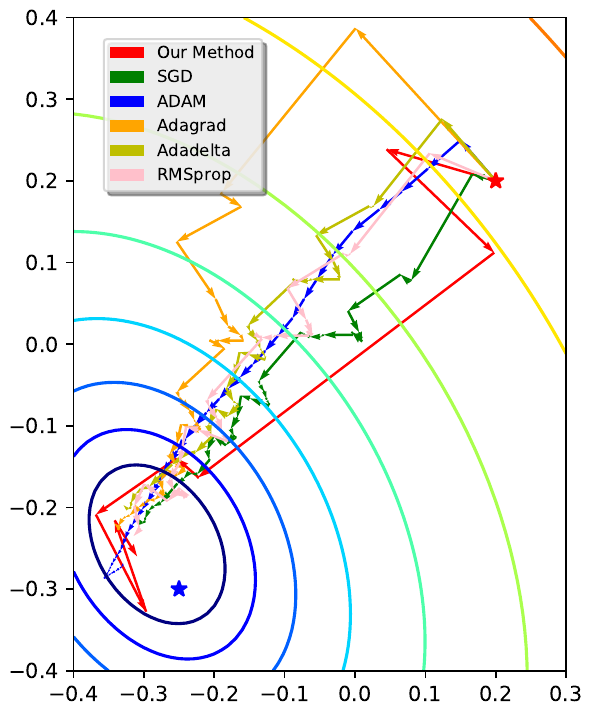}
		
		(c)
	\end{minipage}
	\vspace*{-2mm}
	\caption{Trajectories produced by different algorithms on three random two-dimensional regression problems. The axes represent the values of the two dimensions. The contours outline the area with the same target value, and the target value is gradually decreasing from orange area to blue area. Each arrow represents one iteration of an algorithm, whose tail and tip correspond to the preceding and subsequent iterations respectively.}
	\label{visual}
	\vspace*{-2mm}
\end{figure*}

\subsubsection{Results on CIFAR-10}

CIFAR-10 is a dataset consisting of 60000 natural $32\times32$ RGB images in 10 classes: 50,000 images for training and 10,000 for test. We used a CNN with 2 convolutional layers (each followed by max-pooling layer) and 2 fully connected layers for this task.
Before inputting an image to the CNN, we subtracted the per-pixel mean computed over the training set from each image.

Figure \ref{exp_cifar10} shows the results of all the algorithms on CIFAR-10, including the curves of training loss and test loss. 
While the convergence speed of our method is similar to that of the baselines, the final performance of our method is the best among all the compared algorithms.

\subsection{Further Analysis}
\begin{figure}		
	\centering		
	\includegraphics[page=28, width=0.95
	\linewidth]{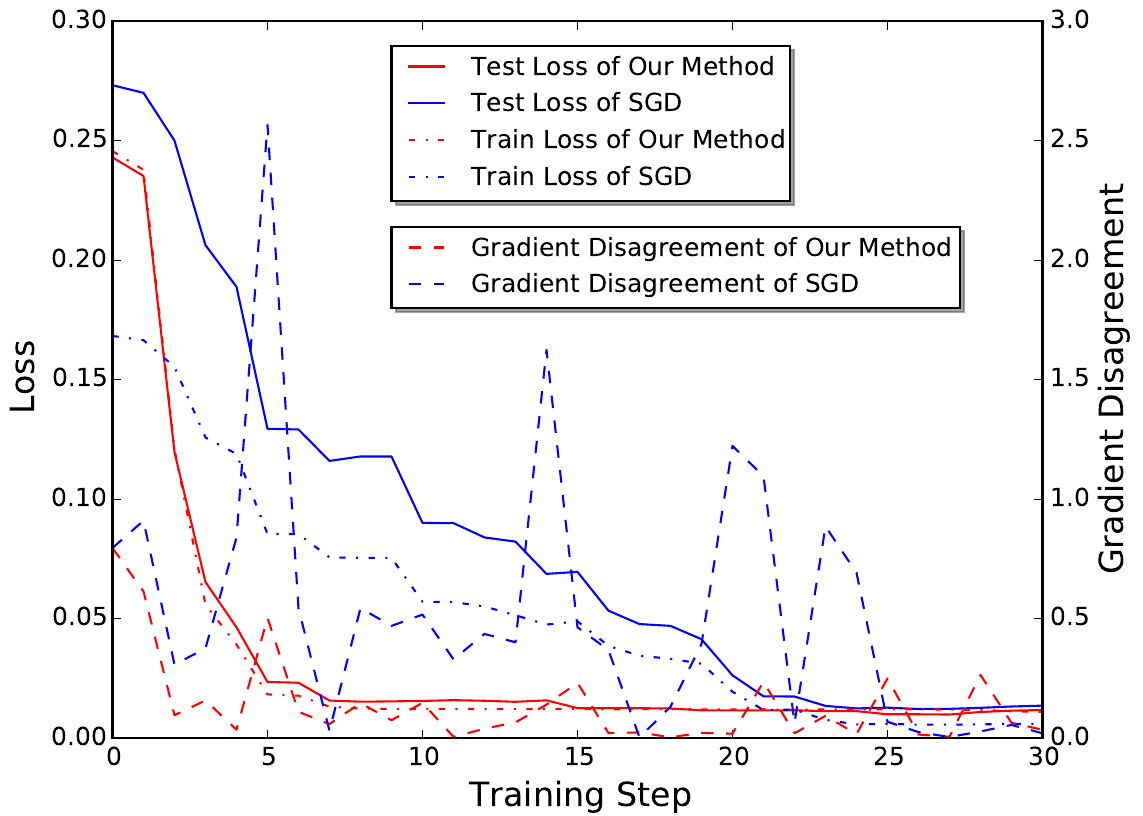}
	\vspace*{-2mm}		
	\caption{Gradient disagreement, training loss and test loss of SGD and our method on a two-dimensional regression problem.}
	\vspace*{-3mm}	
	\label{disagree}
\end{figure}
In order to verify our intuitive explanation that by considering gradient disagreement, our method can make the learning process of the primary ML algorithm stable, here we conducted another experiment.
In this experiment, we investigate the relationship between gradient disagreement and loss in the training process of a simple two-dimensional regression problem. We quantify gradient disagreement by using Euclidean distance between gradient on current batch of data and the overall gradient.

Figure \ref{disagree} shows the gradient disagreement, training loss and test loss of SGD and our method. We can observe the correlation among them from the figure.
As discussed in Section \ref{overfitting_section}, by feeding different samples to actor and critic networks, our method would encourage the learning rate to be small when gradient disagreement is large, so that the oscillation of the training process would be relieved.

It is easy to see from the figure that test loss of our method is stable when there is big gradient disagreement, while the loss of SGD oscillates along with gradient disagreement, leading to slow speed of convergence. The test loss of SGD may increase when gradient disagreement increases, while in overall, our test loss decline in monotonous in the figure. Therefore, we need to feed different training data to the actor network and the critic network to ensure the performance of the algorithm.

To get deeper insight, we visualized the optimization process of our method. From Figure~\ref{visual}, we can find that our method get to convergence with fewer steps and the optimization trajectory is relatively smooth compared to other methods.

\section{Conclusions and Future Work}
\label{conclusion}

In this work, we have studied how to control learning rates for gradient based machine learning methods and proposed an actor-critic algorithm, to help the main network achieve better performance. The experiments on two image classification tasks have shown that our method (1) can successfully adjust learning rate for different datasets and CNN model structures, leading to better convergence, and 2) can reduce oscillation during training. 


For the future work, we will explore the following directions. In this work, we have applied our algorithm to control the learning rates of SGD. We will apply to other variants of SGD methods. We have focused on learning a learning rate for all the model parameters. We will study how to learn an individual learning rate for each parameter. We have considered learning learning rates using RL techniques. We will consider learning other hyperparameters such as step-dependent dropout rates for deep neural networks.

\bibliographystyle{named}
\bibliography{ijcai17}

\end{document}